\pdfoutput=1

\documentclass[11pt]{article}

\usepackage{EACL2023}

\usepackage{times}
\usepackage{latexsym}
\usepackage{graphicx}

\usepackage{array}
\usepackage{mathtools}
\usepackage{subcaption}
\usepackage{pbox}
\usepackage{makecell}
\usepackage{float}
\usepackage{tabularx}
\usepackage{amsfonts}
\usepackage[export]{adjustbox}

\newcolumntype{P}[1]{>{\centering\arraybackslash}p{#1}}

\newcommand\aspace{\hspace{.75em}}
\usepackage[T1]{fontenc}

\usepackage[utf8]{inputenc}

\usepackage{microtype}

\usepackage{inconsolata}

\title{Social Influence Dialogue Systems: A Survey of Datasets and Models For Social Influence Tasks}

\author{
Kushal Chawla$^{1}$\hspace{-0.5mm}\thanks{\hspace{0.6mm} Equal contribution,\hspace{0.6mm}$^{**}$Co-supervised project} \aspace Weiyan Shi$^{2*}$\aspace Jingwen Zhang$^{3**}$\aspace \\
\bf
Gale Lucas$^{1**}$\aspace
Zhou Yu$^{2**}$\aspace
Jonathan Gratch$^{1**}$\aspace
\\
$^{1}$University of Southern California \aspace $^{2}$Columbia University\\$^{3}$University of California Davis\\
{\small
$^{1}$\texttt{\{chawla, lucas, gratch\}@ict.usc.edu} \aspace $^{2}$\texttt{\{ws2634, zy2461\}@columbia.edu} } \\
{\small
$^{3}$\texttt{jwzzhang@ucdavis.edu}}
}

\begin{document}
\maketitle
\begin{abstract}





Dialogue systems capable of social influence such as persuasion, negotiation, and therapy, are essential for extending the use of technology to numerous realistic scenarios. However, existing research primarily focuses on either task-oriented or open-domain scenarios, a categorization that has been inadequate for capturing influence skills systematically. There exists no formal definition or category for dialogue systems with these skills and data-driven efforts in this direction are highly limited. In this work, we formally define and introduce the category of \emph{social influence dialogue systems} that influence users' cognitive and emotional responses, leading to changes in thoughts, opinions, and behaviors through natural conversations. We present a survey of various tasks, datasets, and methods, compiling the progress across seven diverse domains. We discuss the commonalities and differences between the examined systems, identify limitations, and recommend future directions. This study serves as a comprehensive reference for social influence dialogue systems to inspire more dedicated research and discussion in this emerging area.
\end{abstract}

\section{Introduction}

\defcitealias{thomas2006get}{Th06}
\defcitealias{fornaciari2012decour}{Fo12}
\defcitealias{danescu2012echoes}{Da12}
\defcitealias{devault2015toward}{De15}
\defcitealias{asher2016discourse}{As16}
\defcitealias{tan2016winning}{Ta16}
\defcitealias{althoff2016large}{Al16}
\defcitealias{tanana2016comparison}{TaC16}
\defcitealias{lewis2017deal}{Le17}
\defcitealias{yarats2018hierarchical}{Ya18}
\defcitealias{he2018decoupling}{He18}
\defcitealias{zhou2019augmenting}{Zh19}
\defcitealias{tang2019target}{Ta19}
\defcitealias{durmus2019corpus}{Du19}
\defcitealias{demasi2019towards}{De19}
\defcitealias{rashkin2019towards}{Ra19}
\defcitealias{wang2019persuasion}{Wa19}
\defcitealias{zhang2020learning}{Zh20}
\defcitealias{joshi2020dialograph}{Jo20}
\defcitealias{li2020end}{Li20}
\defcitealias{ji2020cross}{Ji20}
\defcitealias{peskov2020takes}{Pe20}
\defcitealias{yang2021improving}{Ya21}
\defcitealias{liu2021towards}{Li21}
\defcitealias{liang2021evaluation}{LiE21}
\defcitealias{yamaguchi2021dialogue}{YaD21}
\defcitealias{hadfi2021argumentative}{Ha21}
\defcitealias{boritchev2021ding}{Bo21}
\defcitealias{wu2021alternating}{Wu21}
\defcitealias{jhan2021cheerbots}{Jh21}
\defcitealias{chawla2021casino}{Ch21}

Consider a human user who signs up to interact with a persuasive dialogue system that motivates for engaging in physical exercise. The system: 1) uses \textit{social cues like small talk and empathy}, useful for providing continued support, and 2) employs \textit{persuasive strategies to convince} the user who, at least in the short-term, is reluctant to indulge in exercise.
Does such a system fit the definition of a \textit{task-oriented dialogue system} that are traditionally designed to assist users in completing their tasks such as restaurant or flight booking~\cite{zhang2020recent}? Although the system is task-oriented or goal-oriented per se, the task here goes beyond the traditional definition of assisting users, given the possible misalignment between the goals of the system and the user. Clearly, this system is also not open-domain~\cite{huang2020challenges}. Although conversations involve social open-ended interactions, there is still a concrete goal of persuading the user towards a healthier habit.

Scenarios similar to above are ubiquitous in everyday life, including games~\cite{peskov2020takes}, social platforms~\cite{tan2016winning}, and therapeutic interactions~\cite{tanana2016comparison}. Dialogue systems for these applications require a core function in human communication, that is, \textit{social influence}~\cite{cialdini2004social, cialdini2009influence}, which involves influencing users' cognitive and emotional responses, leading to changes in thoughts, opinions, and behaviors through natural conversations. This goes beyond what is captured by traditional task definitions in the dialogue community which primarily focus on task completion and social companionship.

Despite numerous independent efforts in identifying and analyzing various social influence scenarios, there is a lack of common understanding around social influence in AI research which inhibits a systematic study in this space. Further, data-driven efforts for dialogue systems in this space are highly limited. To this end, we introduce the category of \emph{social influence dialogue systems} (Section~\ref{sec:Social Influence Dialogue Tasks}), providing a comprehensive literature review and discussing future directions.

Developing these systems holds importance in AI research for multiple reasons. Tackling these tasks not only involves AI but also aspects of game theory, communication, linguistics, and social psychology, making them an ideal testbed for interdisciplinary AI research. Most importantly, they reflect AI's general ability to consider their partners' inputs, tailor the communication strategies, personalize the responses, and lead the conversation actively.

We design a taxonomy for existing social influence dialogue datasets, studying their task structure (symmetric vs asymmetric) and context (local vs global). We also organize them by their domains: games, multi-issue bargaining, social good, e-commerce, therapy and support, argumentation, conversational recommendations, and miscellaneous tasks (Section \ref{sec:Social Influence Across Disciplines}). We further design a taxonomy of existing methods, assisting readers to comprehend the progress and reflect on future directions. We organize them based on the system strategy, language generation, partner model, architecture, learning process, and the use of pretrained language models (Section \ref{sec:methods}). Finally, we identify key challenges and provide recommendations for future work (Section~\ref{sec:recommendations}).

Over the years, research in task-oriented and open-domain dialogues has benefited from a myriad of survey efforts~\cite{huang2020challenges,zhang2020recent,ni2021recent}. We instead focus on dialogue systems with social influence capabilities and present a thorough review across various domains. We hope that our work serves as a timely entry point for interested researchers to take this area further, inspiring dedicated effort and discussion on social influence in the dialogue community.

\section{Social Influence Dialogue Systems}
\label{sec:Social Influence Dialogue Tasks}

``Social influence is a fact of everyday life''~\cite{gass2015social}. It is the \textit{change in thoughts, feelings, attitudes, or behaviors resulting from interaction with an individual or a group}~\cite{rashotte2007social}. Influence is measured by quantifiable proxies of the observed change, like the interest to indulge in physical exercise before or after the interaction with a system, or the final deal in a negotiation as opposed to one person taking it all. Social influence dialogue systems act interactively and influence their partners in decision-making and behavioral contexts~\cite{zhang2020artificial, lee2020designing}. This calls for an active role by the system, distinguishing them from other well-studied scenarios, such as purely task-oriented, where systems passively assist their partners to complete tasks, and open-domain, that target social companionship. Key social influence tasks include persuasion \cite{wang2019persuasion}, aiming to change users' attitudes or behaviors, and negotiation, aiming to change the users' perspective to achieve a common ground \cite{lewis2017deal}.

\begin{figure}[t!]
\centering
 \includegraphics[width=\linewidth]{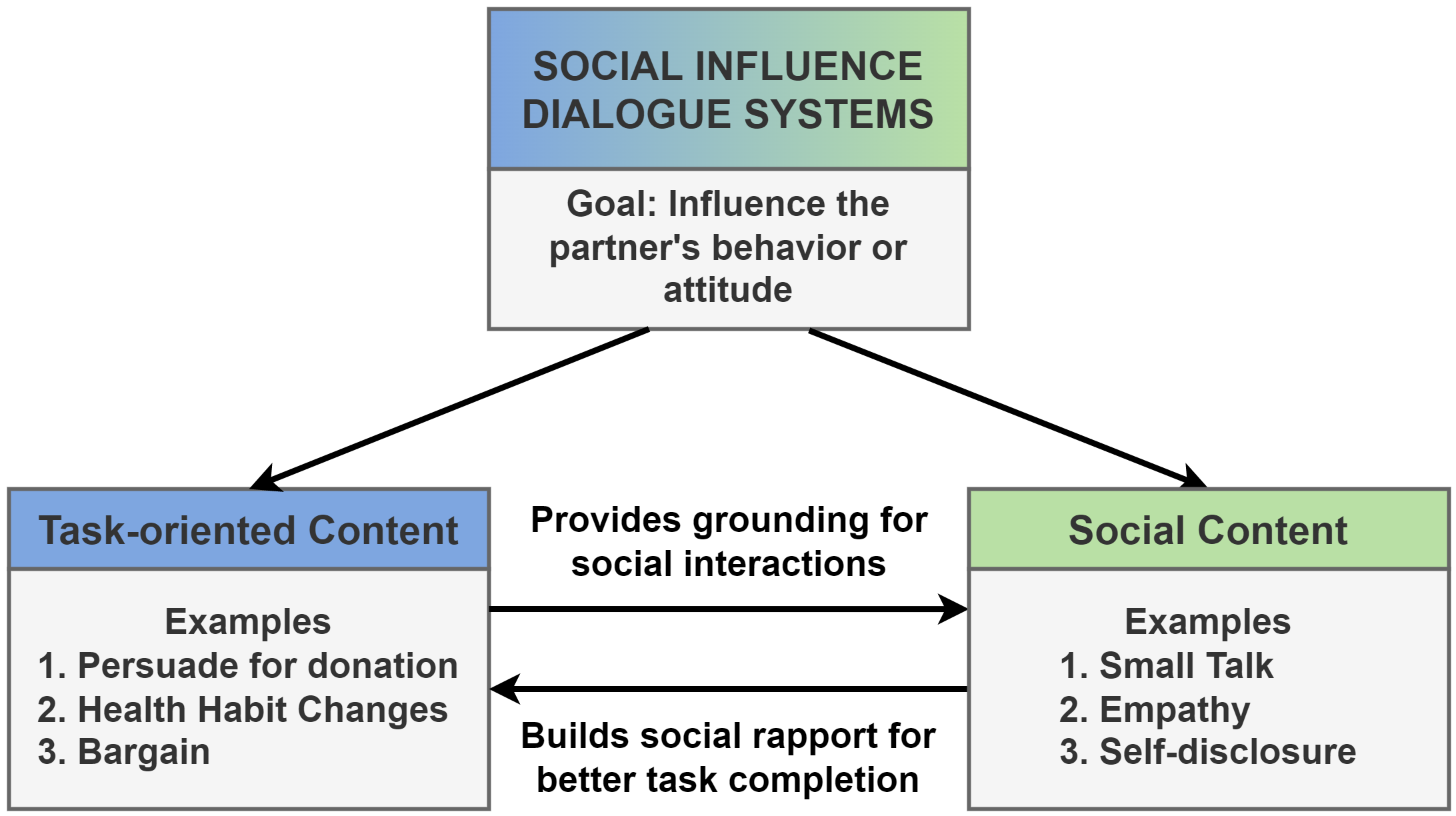}
\caption{A conceptual overview.}
\label{fig:conceptual}
\end{figure}

\noindent\textbf{Conceptual overview}: Figure~\ref{fig:conceptual} distinguishes between the kinds of conversational content in social influence interactions. The \textit{task-oriented content} focuses on influencing for a domain-specific goal, like persuading for donation, bargaining with trade-offs, or encouraging healthier habits. These interactions may also contain \textit{social content}, such as small talk, empathy, or self-disclosure. The task-oriented content provides a context for social interactions. Depending on the task, social content is optional, but if present, can in turn build rapport and enhance user-system relationship for improved task outcomes~\cite{liao2021linguistic}.


\noindent\textbf{Connections with task-oriented and open-domain systems}: Similar to a task-oriented or an open-domain scenario, social influence dialogue can also be seen as a sequential decision making process with the goal of maximizing the expected reward~\cite{huang2020challenges,gao2018neural}. Our proposed category is not meant to be disjoint from these traditional categories. However, it still uniquely brings together the tasks that capture social influence, \textit{which is fundamentally absent from how we primarily define dialogue tasks in the community}. Defining a new category that captures social influence dialogue would foster a dedicated effort towards this important aspect of real-world conversations.

Task-oriented scenarios focus on collaborative information exchange for a common goal of task completion. In social influence tasks, the goals of the system and the user can be different and even conflicting, leading to collaborative or non-collaborative interactions. Further, the goals can go beyond the current task (e.g. multiple therapy interactions, repeated negotiations), leading to social interactions for long-term relationships. If a scenario involves the system's goal to influence its partner, we consider it under social influence in this paper. For instance, \citet{he2018decoupling} studied buyer-seller price negotiations. The task of the buyer is to negotiate for a reasonable price (arguably making it task-oriented), but achieving it requires social influence skills of engaging in trade-offs and building a rapport with the seller so as to reach an agreement. 





\noindent\textbf{Measures of Success}: The above discussion indicates that a comprehensive evaluation of social influence systems must draw from both task-oriented and open-domain dialogue research. Since there exist surveys that discuss the evaluation in these settings~\cite{deriu2021survey,li2021evaluate}, we don't cover them here in detail. However, we define three essential axes for evaluation: 1) \textit{Linguistic Performance}, or the system's linguistic sophistication based on automatic (e.g. perplexity, BLEU) and human (e.g. fluency, consistency, coherency) evaluation. 2) \textit{Influence Outcome}, or the ability to influence defined by objective goals like the negotiated price or weight loss after therapy. 3) \textit{Partner Perception}, or the subjective evaluation of the user, for instance, the user's satisfaction, likeness towards the system, and interest in interacting again. In a buyer-seller negotiation, if the seller hates the buyer in the end, no matter how favorable the deal is for the buyer, one might argue that this is still a failed negotiation for the buyer. Hence, we encourage future work to take all three dimensions into account collectively.

\section{Social Influence Across Diverse Application Areas}
\label{sec:Social Influence Across Disciplines}


\begin{table*}[th!]
\centering
\scalebox{0.76}{
\begin{tabular}{lcP{5cm}ccc}
\hline
\multicolumn{1}{c}{\textbf{Name (Citation)}} & \textbf{Domain} & \textbf{Source} & \textbf{Structure} & \textbf{Context} & \textbf{\# of Parties} \\ \hline
\makecell[l]{\textbf{STAC} (\citetalias{asher2016discourse})} & Games & Crowdsource & Symmetric & Global & Multiparty \\
\makecell[l]{\textbf{Diplomacy} (\citetalias{peskov2020takes})} &  Games & Crowdsource &Asymmetric & Global & Multiparty \\
\makecell[l]{\textbf{DinG} (\citetalias{boritchev2021ding})} & Games & University game night logs &  Symmetric & Global & Multiparty \\
\makecell[l]{\textbf{Tabletop} (\citetalias{devault2015toward})} &  MIBT & Face-to-face, Wizard-of-Oz & Symmetric & Local & Bilateral \\
\makecell[l]{\textbf{DealOrNoDeal} (\citetalias{lewis2017deal})} & MIBT & Crowdsource & Symmetric & Local & Bilateral \\
\makecell[l]{\textbf{CaSiNo} (\citetalias{chawla2021casino})} &  MIBT &Crowdsource &Symmetric & Local & Bilateral \\
\makecell[l]{\textbf{JobInterview} (\citetalias{yamaguchi2021dialogue})} &  MIBT & Crowdsource & Asymmetric & Local & Bilateral \\
\makecell[l]{\textbf{PersuasionforGood} (\citetalias{wang2019persuasion})} & Social Good & Crowdsource &Asymmetric & Global & Bilateral \\
\makecell[l]{\textbf{CraigslistBargain} (\citetalias{he2018decoupling})} & E-commerce & Crowdsource &Asymmetric & Local & Bilateral \\
\makecell[l]{\textbf{AntiScam} (\citetalias{li2020end})} &  E-commerce & Crowdsource & Asymmetric & Global & Bilateral \\
\makecell[l]{\textbf{MI} (\citetalias{tanana2016comparison})} &  Therapy \& Support & Psychotherapy session logs & Asymmetric & Global & Bilateral \\
\makecell[l]{\textbf{SMS Counseling} (\citetalias{althoff2016large})} &  Therapy \& Support & SMS chat logs &Asymmetric & Global & Bilateral \\
\makecell[l]{\textbf{EmpatheticDialogues} (\citetalias{rashkin2019towards})} & Therapy \& Support & Crowdsource & Asymmetric & Global & Bilateral \\
\makecell[l]{\textbf{Hotline Counseling} (\citetalias{demasi2019towards})} & Therapy \& Support & Synthetic Transcripts & Asymmetric & Global & Bilateral \\
\makecell[l]{\textbf{mPED} (\citetalias{liang2021evaluation})} & Therapy \& Support & Physical activity clinical trials &Asymmetric & Global & Bilateral \\
\makecell[l]{\textbf{Congressional Debates} (\citetalias{thomas2006get})} &  Argumentation & U.S. Congressional transcripts &Asymmetric & Local & Multiparty \\ 
\makecell[l]{\textbf{Supreme Court} (\citetalias{danescu2012echoes})} &  Argumentation & Oyez.org transcripts &Asymmetric & Local & Multiparty \\ 
\makecell[l]{\textbf{DeCour} (\citetalias{fornaciari2012decour})} & Argumentation & Italian court hearings &Asymmetric & Local & Multiparty \\ 
\makecell[l]{\textbf{ChangeMyView} (\citetalias{tan2016winning})} & Argumentation & Reddit & Asymmetric & Local & Multiparty \\ 
\makecell[l]{\textbf{DDO Debates} (\citetalias{durmus2019corpus})} & Argumentation & debate.org logs & Symmetric & Local & Bilateral \\
\makecell[l]{\textbf{Court Debates} (\citetalias{ji2020cross})} &  Argumentation & China Court transcripts &Asymmetric & Local & Multiparty \\ 
\makecell[l]{\textbf{Target-Guided} (\citetalias{tang2019target})} & Miscellaneous & Crowdsource &Symmetric & Local & Bilateral \\ \hline
\end{tabular}}
\caption{\label{tab:corpora-categories} Categorization of social influence dialogue corpora. This list is non-exhaustive, and also covers the datasets that have enabled research into various sub-tasks and analyses that can eventually be useful for dialogue systems in respective domains. MIBT: Multi-Issue Bargaining Task. Key statistics and associated metadata are in Appendix \ref{tab:corpora}.}
\end{table*}


We now illustrate social influence across numerous domains and application areas. In total, we curated $22$ datasets from prior work that capture social influence in various forms, spanning $12$ publication venues, $4$ languages, and $7$ application domains (see Appendix \ref{sec:appendix-sources} for details on the compilation process). In general, the datasets capture the following information about an interaction: the \textit{non-conversational context} for the participants (e.g. negotiation preferences or other role-specific information), the \textit{conversation} between them, and \textit{outcome assessment}. Optionally, some datasets also gather participant demographics and personality traits, utterance-level annotations, and subjective evaluations via post-surveys.

To understand the structural similarities and differences between these datasets, we design a taxonomy with two primary dimensions: \textbf{Task Structure} (\textit{Symmetric} vs \textit{Asymmetric}), and \textbf{Context Definition} (\textit{Global} vs \textit{Local}). \textbf{Task Structure} captures whether the participant roles are defined in a symmetric or an asymmetric manner. For instance, a typical multi-issue negotiation is symmetric, in the sense that both parties have their own preferences and goals based on which they actively try to reach a favorable agreement~\cite{lewis2017deal}. On the other hand, a counseling session between a therapist and a patient is asymmetric, where the therapist attempts to emotionally support the patient by employing social influence skills~\cite{althoff2016large}. \textbf{Context Definition} relates to whether the input context before each interaction is defined globally or locally. For instance, the PersuasionForGood dataset globally defines the context of persuasion for charity donation, which is kept the same throughout~\cite{wang2019persuasion}. On the contrary, in a typical debate, although the rules are defined globally, the conversation topic and arguments are local and can vary for each conversation~\cite{durmus2019corpus}. We present this categorization in Table \ref{tab:corpora-categories}. We further categorize the datasets according to their \textbf{Domain}, \textbf{Source}, and the \textbf{\# of parties}. We provide key statistics and the available metadata in Appendix \ref{sec:appendix-datasets}. We now briefly discuss the datasets in each domain.


\noindent\textbf{Games}: Strategy games involve social influence dynamics of trust and deception. Diplomacy captures deception in long-lasting relationships, where players forge and break alliances to dominate Europe~\cite{peskov2020takes}. Catan revolves around the trade of resources for acquiring roads, settlements, and cities~\cite{asher2016discourse,boritchev2021ding}. The players have access to only a subset of resources that they would need, which encourages strategic influence and trade.

\noindent\textbf{Multi-Issue Bargaining Tasks (MIBT)}: MIBT is a tractable closed-domain abstraction of a typical negotiation~\cite{fershtman1990importance}. It is based on a fixed set of issues each with a predefined priority for each player, which essentially governs the goals of the players. If the priorities of the players align, this leads to competitive negotiations, where each party attempts to convince their partner with trade-offs and persuasive arguments. If they don't, this allows cooperative interactions where the negotiators try to find optimal divisions that benefit everyone. DealOrNoDeal~\cite{lewis2017deal} involves negotiations over three issues: \textit{books}, \textit{balls}, and \textit{hats}. Other datasets define a more grounded scenario, such as symmetric CaSiNo~\cite{chawla2021casino} negotiations between two campsite neighbors and asymmetric JobInterview~\cite{yamaguchi2021dialogue} negotiations between recruiters and applicants.

\noindent\textbf{Social Good}: Social influence is critical for social good applications. The tactics must be personalized using knowledge that is both relevant and appealing. PersuasionForGood~\cite{wang2019persuasion} involves asymmetric interactions led by a persuader who attempts to convince the other participant for charity donation by employing a variety of tactics. For instance, \textit{Logical Appeal} uses reason and evidence to support the argument, while \textit{Emotional Appeal} elicits specific emotions.

\noindent\textbf{E-commerce}: These tasks are typically asymmetric. A \textit{buyer} influences the seller towards a reasonable price, while the \textit{seller} tries to maximize their own profit. An effective system must combine price-related reasoning with language realization. CraigslistBargain~\cite{he2018decoupling} involves open-ended price negotiations with rich influence strategies like embellishments, side offers, emotional appeals, and using world knowledge. Another example is customer support interactions in AntiScam dataset~\cite{li2020end}, where users defend themselves against attackers who try to steal sensitive personal information with convincing arguments.

\noindent\textbf{Therapy \& Support}: Effective therapy using social influence aids in the treatment of mental disorders, and substance use disorders, along with changing undesirable behaviors like unhealthy diets. A counselor needs to be adaptive, personalized, should understand the core issues, and should facilitate a change in patient's perspective~\cite{althoff2016large}. In SMS counseling,~\citet{althoff2016large} found that linguistic influence like pushing the conversation in the desired direction is associated with perspective change. Similar scenarios were captured in other datasets as well~\cite{demasi2019towards,liang2021evaluation}.~\citet{tanana2016comparison} collected the Motivational Interviewing dataset where the goal is to elicit and explore the patient's own motivations for behavior change. EmpatheticDialogues~\cite{rashkin2019towards} captured empathetic support interactions, which has been associated with rapport and better task outcomes~\cite{kim2004effects,norfolk2007role,fraser2018spoken}.


\noindent\textbf{Argumentation}: 
In addition to factuality and social proof, a convincing argument must also consider the intensity, valence, authoritativeness, and framing~\cite{chaiken1987heuristic,althoff2014ask}.~\citet{tan2016winning} released the ChangeMyView logs from Reddit, involving discussions on numerous controversial topics. Other datasets include Debate Dot Org (DDO) debates on diverse topics~\cite{durmus2019corpus}, congressional proceedings~\cite{thomas2006get}, and court hearings~\cite{fornaciari2012decour,danescu2012echoes,ji2020cross}.

\noindent\textbf{Conversational Recommendation}: Everyday scenarios naturally hold potential for influence via recommendations, for instance, a movie fan persuading their friends to watch a movie that they adore.~\citet{li2018towards} and ~\citet{dodge2016evaluating} collected movie recommendation datasets. Instead of guiding the conversation towards a specific movie, the goal is simply to provide recommendations based on facts and personal experiences. Nevertheless, they still provide interesting examples of scenarios that can involve social influence.

\noindent\textbf{Miscellaneous}: The Target-Guided dataset~\cite{tang2019target} was constructed from the PersonaChat corpus~\cite{zhang2018personalizing}. Instead of being open-ended, the Target-Guided scenario defines a concrete goal of naturally guiding the conversation to a designated target subject, thereby, making it a social influence setting.

\section{Methodological Progress}
\label{sec:methods}

\begin{table*}[th!]
\centering
\begin{adjustbox}{width=.99\textwidth}
\begin{tabular}{lccccccc}
\hline
\multicolumn{1}{c}{\textbf{Method}} & \textbf{Domain} & \textbf{Strategy} & \textbf{NLG} & \textbf{Partner Model} & \textbf{Architecture} & \textbf{Learning} & \textbf{PLM} \\ \hline
\citetalias{lewis2017deal} & MIBT & Implicit & Generation & Simulated User & Enc-Dec & RL & \\
\citetalias{yarats2018hierarchical} & MIBT & Latent Vectors & Generation & Simulated User & Hierarchical & RL & \\
\citetalias{zhang2020learning} & MIBT & DAs & Generation & DA Look-Ahead & Modular & RL & \\
\citetalias{he2018decoupling} & E-Com, MIBT & DA & Templates + Retrieval & Implicit &  Modular & RL & \\
\citetalias{yang2021improving} & E-Com & DAs & Templates + Retrieval & DA Look-Ahead & Modular & RL &  \\
\citetalias{zhou2019augmenting} & E-Com, Social Good & DAs + Semantic & Generation & Implcit & Hierarchical & SL & \\
\citetalias{joshi2020dialograph} & E-Com &  DAs + Semantic & Generation & Implcit & Hierarchical & SL & \\
\citetalias{li2020end} & E-Com, Social Good & DAs & Generation & Implicit & Dec-only & SL & GPT\\
\citetalias{liu2021towards} & Therapy & Implicit & Generation & Implicit & Enc-Dec & SL & \\
\citetalias{jhan2021cheerbots} & Therapy & Emotion Labels & Retrieval, Generation & Simulated User & Modular & RL & BERT, GPT \\
\citetalias{hadfi2021argumentative} & Argumentation & DAs & Rule-based & Implicit & Modular & SL & \\
\citetalias{wu2021alternating} & Social Good & Implicit & Generation & Implicit & Dec-only & SL & GPT2\\
\citetalias{tang2019target} & Misc & Keywords & Retrieval & Implicit & Modular & SL & \\ \hline
\end{tabular}
\end{adjustbox}
\caption{\label{tab:methods} Categorization of methods (non-exhaustive) for social influence dialogue. We only cover papers that explicitly design a dialogue system. NLG: Natural Language Generation, PLM: Pretrained Language Model, MIBT: Multi-Issue Bargaining Task, E-Com: E-Commerce, DA: Dialogue Act, Enc: Encoder, Dec: Decoder, SL: Supervised Learning, RL: Reinforcement Learning. Methods that use RL usually apply it in conjunction with SL.}
\end{table*}

Having summarized the datasets that capture social influence, we now discuss the modeling approaches developed for social influence dialogue systems. Most domains have seen efforts in analyzing human dialogue behaviors and their impact on task outcomes. Examples include analyzing deception in games~\cite{peskov2020takes}, the impact of persuasive strategies and dialogue acts on charity donations~\cite{wang2019persuasion}, cooperative and non-cooperative strategies in MIBT~\cite{chawla2021casino}, the use of emotion expression for predicting partner perceptions~\cite{chawla2021towards}, and studying semantic categories of persuasive arguments on web forums~\cite{egawa2019annotating}.

In addition, researchers have targeted various domain-specific subtasks that can be crucial for the eventual development of dialogue systems in this space. This involves research in lie detection methods~\cite{yeh2021lying,yu2015detecting}, discourse parsing~\cite{shi2019deep,ouyang2021dialogue}, strategy prediction~\cite{chawla2021casino,wang2019persuasion}, breakdown detection~\cite{yamaguchi2021dialogue}, outcome prediction~\cite{sinha2021predicting,chawla2020exploring,dutt2020keeping}, and argument mining~\cite{dutta2022can}.


Research that directly targets the development of dialogue systems in this space is still nascent. Among other challenges like limited cross-cultural diversity and relatively smaller dataset size, social influence dialogue settings pose a unique challenge: an average human often exhibits sub-optimal strategic behaviors in social influence tasks~\cite{wunderle2007negotiate, babcock2009women}. This means that standard seq2seq approaches trained on these collected datasets using supervised learning are fundamentally insufficient for developing effective dialogue systems with influence capabilities. Hence, prior work has put a special attention to the system strategy, employing different ways to model the strategy and language together.

We design a taxonomy of methods developed for social influence tasks, assisting readers to comprehend the progress and reflect on future directions. We organize them based on the system strategy, language generation, partner model, architecture, learning process, and the use of pretrained language models. We present annotations for all the surveyed methods in Table \ref{tab:methods} and discuss the common categories in brief below.

\subsection{Strategy Representation}
\noindent\textbf{Implicit}: The most obvious way to represent the system strategy is \textit{implicitly}, without any intended decoupling between system strategy and response realization. This corresponds to the usual sequence-to-sequence framework that has been a standard baseline for the methods developed in this space. An important example is the work by \citet{lewis2017deal}, who were one of the first works to train end-to-end dialogue models that exhibit social influence. The authors employed a neural network based on GRUs, one for encoding the negotiation context, one to encode the dialogue utterances, and two recurrent units to generate the output agreement in a bidirectional manner.

\noindent\textbf{Latent vectors}:~\citet{yarats2018hierarchical} explored \textit{latent vectors} to decouple utterance semantics from its linguistic aspects. Their hierarchical approach first constructs a latent vector from the input message, which is then used for response generation and planning. These latent vectors are trained to maximize the likelihood of future dialogue messages and actions, which enables the decoupling between semantics and realization.

\noindent\textbf{Dialogue Acts (DAs)}: Dialogue Acts, such as greeting, offer propose, agreement, or disagreement, are effective at capturing a high-level structure of the dialogue flow in social influence settings, reducing the model strategy to first predicting the dialogue act for the next response. The use of DAs makes it convenient to apply reinforcement learning approaches~\cite{zhang2020learning,yang2021improving}, while also aiding in developing a modular dialogue system design~\cite{he2018decoupling}.

\noindent\textbf{Semantic Strategies}: The structural properties expressed by DAs are insufficient for capturing semantics like emotion, small talk, and appeal. To better incorporate them, researchers have relied on additional utterance-level annotations grounded in prior theories in social influence contexts~\cite{wang2019persuasion,chawla2021casino}. These strategies have been used in conjunction with DAs~\cite{zhou2019augmenting,joshi2020dialograph}.

\subsection{Language Generation}
An important aspect of the system design is an effective way to realize the language, that is, to generate the next response so that it portrays the desired strategic behaviors. Borrowing from task-oriented and open-domain research, existing dialogue models for social influence use a variety of methods to generate the final system response.

\noindent\textbf{Templates and retrieval methods}: Predefined templates and response retrieval from the training data simplify the generation pipeline, improving controllability and modularity.~\citet{he2018decoupling} used templates in their generator which are later filled by retrieving similar responses from the data. This allowed the authors to explore supervised and reinforcement learning at the level of DAs for the influence strategy of the system.

\noindent\textbf{Conditional Generation}: Text generation methods result in more diverse responses, but negatively impact the controllability and interpretability. Prior work relies on autoregressive text generation conditioned on the dialogue history, non-conversational context, and additional annotations. These are either encoder-decoder networks~\cite{lewis2017deal,li2020end,joshi2020dialograph} or use a decoder-only design~\cite{li2020end}. A useful future direction is to combine generation with retrieval for knowledge-grounded settings like argumentation. Similar methods have been explored for other NLP tasks like open-domain question answering and question generation~\cite{lewis2020retrieval}.

\subsection{Partner Modeling}
\textit{Partner modeling} refers to inferring the mental states of the partner based on the conversation. For example, understanding the cause that the persuadee cares about in the PersuasionForGood context, or inferring the priorities of the partner in DealOrNoDeal negotiations. Building an accurate partner model is essential in social influence settings for guiding the decision-making of the system~\cite{baarslag2013predicting,zhang2020learning}. Hence, we discuss various ways in which prior work tackles partner modeling.

\noindent\textbf{Implicit}: A majority of the efforts do not explicitly model the behavior of the partner but instead, this behavior implicitly guides the next response of the sequence-to-sequence dialogue system pipeline.

\noindent\textbf{Simulated User}:~\citet{lewis2017deal} trained a simulated user on the available data in a supervised manner. This was then used to further train the dialogue system. Instead of inferring mental states explicitly, this takes a more behavioral approach of estimating the future actions of the partner and using these for training via reinforcement learning.

\noindent\textbf{Dialogue Act Look-Ahead}: With a similar idea,~\citet{zhang2020learning} proposed OPPA model with a \textit{look-ahead} based partner modeling strategy at the level of DAs. At each step, OPPA first estimates the user's future DA, which is then used to select the next DA of the system. The authors found significant improvements on the DealOrNoDeal task.~\citet{yang2021improving} used a similar method for buyer-seller negotiations. Taking a different approach,~\citet{chawla2022opponent} trained a ranking model to directly predict the hidden preferences of the partner in a multi-issue negotiation. Instead of predicting future actions, these methods assume that the partner's behavior can be explained by their context and goals in the dialogue. However, this approach is yet to be used in an end-to-end system.

\subsection{Training}
\noindent\textbf{Architecture Choices}: One crucial aspect is the \textit{architecture design}: End-to-end~\cite{lewis2017deal,radford2019language} vs Modular~\cite{he2018decoupling}. While end-to-end methods improve the diversity and need less manual effort, a modularized design enhances controllability and explainability. Perhaps, this is why modular methods are popular in large-scale models~\cite{hadfi2021argumentative}. Improving the control of desired variables such as topics, strategy, or emotion in the end-to-end methods is an open area of research and is yet to be explored for social influence dialogue systems.

\noindent\textbf{Supervised Learning (SL) and Reinforcement Learning (RL)}: ~\citet{zhou2019augmenting} used SL to train a hierarchical encoder-decoder for generating the next response and used Finite State Transducers (FSTs) to encode the historic sequence of DAs and persuasive strategies into the model, showing improvements in negotiation and persuasion tasks. The performance was later improved by~\citet{joshi2020dialograph}, who replaced FSTs with Graph Neural Networks to better model the interdependencies. Others have relied on RL to explicitly optimize the model on task-specific objective outcomes. While SL trains the model to mimic the average human behavior, RL techniques, such as those based on REINFORCE~\cite{williams1992simple}, allow the system to explore its own strategies in the wild while being guided by one or more overall reward metrics.~\citet{lewis2017deal} used RL in negotiations, with the final points scored in the agreed deal as the reward. More recent work employed RL to incorporate simplistic partner models into the decision-making process of the dialogue system, showing improvements in negotiation tasks~\citep{zhang2020learning,yang2021improving}.

\noindent\textbf{Multi-tasking and Pretraining}: Limited efforts have also explored multi-tasking and pretrained language models for social influence dialogue systems, which provide promising ways to deal with the challenge of insufficient training data.~\citet{liu2021towards} trained a sequence-to-sequence transformer on a mix of Cornell Movie Dialogue corpus~\cite{danescu2011chameleons} and psychotherapy data.~\citet{li2020end} fine-tuned the GPT model~\cite{radford2018improving}, while employing multi-tasking to incorporate intents and slots for both the human and the system.~\citet{wu2021alternating} recently introduced ARDM which uses GPT2~\cite{radford2019language} to separately encode the utterances of the human and the dialogue system, reducing the reliance on additional annotations.


\section{Discussion and Recommendations}
\label{sec:recommendations}

Past few years have seen an exciting progress in social influence dialogue systems. However, building sophisticated and practically useful systems remains a challenging endeavor. Several limitations still exist that must be addressed. To guide future work, we now discuss the key challenges and provide our recommendations.

\noindent\textbf{Need for unifying the efforts}: One challenge in this space has been the lack of large-scale datasets for model training. Social influence tasks are complex for crowdsourcing workers to understand and to participate in. Hence, prior work used extensive instructions and tutorials, making the study expensive and time consuming~\cite{wang2019persuasion,chawla2021casino}. To address this, we recommend the researchers to aim for a more unified view of the efforts in social influence.

First, this would encourage researchers to adopt the best practices from other social influence scenarios. For instance, most datasets miss out on user attributes like demographics and personality, which are crucial in social influence scenarios~\cite{stuhlmacher1999gender,bogaert2008social}. Most datasets also ignore the partner perception after the interaction is over. This can result in misleading conclusions about the model performance, where the models perform well objectively, but hurt the relationship with their partners, and thus, negatively impacting practical utility~\cite{aydougan2020challenges}.

Secondly, a holistic outlook will promote transfer learning and domain adaptation. Our taxonomy for datasets (Table \ref{tab:corpora-categories}) governs the way systems must be modeled and trained. Task structure is crucial to understand whether the model can learn from the utterances of all parties or just one. Further, understanding the context definition guides how it must be encoded. Hence, one interesting future direction is joint training on datasets with similar structure and context definition.

Finally, progress in task-oriented and open-domain systems can inspire more unified modeling for social influence tasks involving multiple skills in the same interaction (e.g. a combination of negotiation and persuasion tactics as common in realistic scenarios).~\citet{roller2020recipes} blend various open-domain tasks to address multiple challenges together (e.g., persona-based, knowledge-enriched, etc.). \citet{hosseini2020simple} concatenate structured and unstructured data in task-oriented dialogues, and unify task-oriented dialogue system building to be a single sequence generation problem. Future work should explore similar unified approaches for social influence settings as well, especially since these tasks follow a common conceptual foundation (Figure \ref{fig:conceptual}), with similar evaluation and theoretical principles~\cite{cialdini2009influence}.

To encourage this unified view, we encapsulate our insights from this survey effort in a theoretical framework, which is presented in Appendix \ref{sec:appendix-framework}. The framework covers key components for designing a social influence dialogue task, including system attributes, target audience, underlying modeling techniques, and evaluation mechanisms.

\noindent\textbf{Theory integration}: Most modeling efforts are based on crowdsourced datasets. Since crowdsourcing workers may not exhibit optimal strategies, supervised training on these datasets is fundamentally insufficient to build an effective system for applications like pedagogy (teaching social skills to students). Unfortunately, this holds regardless of how system strategy and partner model are designed. Further, using RL to optimize on objective rewards is also not expected to be enough to reliably learn complex influence capabilities, especially when the reward is restrictive.

To address this, we recommend to tap into the vast amount of research effort in social sciences and psychology on building theories for social influence~\cite{cameron2009practitioner,giles2016communication,lewicki2016essentials,cialdini2004social}. Instead of solely relying on the collected data, future work should consider leveraging fundamentals from this research to guide the dialogue policy. Previous works have studied resistance to social influence \cite{knowles2004resistance, dal2004narrative, petty1977forewarning, ahluwalia2000examination}. \citet{rucker2004individual} found that people resist persuasion differently depending on their beliefs, suggesting personalizing the social influence process. One can also employ the politeness theory~\cite{brown1978universals} and model the participants' \emph{face} acts to better understand users in social influence contexts~\cite{dutt2020keeping}.

\noindent\textbf{Task Evaluation}: Another key limitation of existing work is the lack of a comprehensive evaluation. Prior work majorly focused on objective metrics which only provides a limited view of the model performance. A comprehensive evaluation is challenging since it must consider partner perception along with objective outcomes. Building user simulators could potentially alleviate this problem~\cite{li2016user, jain2018user, shi2019build}. Most existing simulators are developed for task-oriented systems which follow a certain agenda. Future research should study how to use partner modeling to build social influence user simulators for more efficient and accurate task evaluation \cite{he2018decoupling,yang2020improving}. For instance, one could potentially design different user personalities and simulate the change in user's beliefs, opinions, and attitudes accordingly~\cite{yang2021improving}.

\noindent\textbf{Multimodal systems}: Being a core function of human communication, social influence occurs not just through text, but through all possible modalities. \citet{schulman2009persuading} showed that embodied agents achieve better persuasion results than text-only agents. Other studies have recognized the importance of emotion in social influence tasks~\cite{asai2020emotional,chawla2021towards}. ~\citet{nguyen2021acoustic} proposed a speech dataset in debates and study the influence of spoken tactics on persuasiveness across genders. Given these findings, we encourage interdisciplinary efforts in the future to explore the developement of multimodal social influence agents.

\noindent\textbf{Knowledge-enriched systems}: Social influence tasks often involve constantly-changing world knowledge such as organization facts and news. Often, the system's internal state (e.g., the change of task setting from one set of products to a different set) needs to be updated. Retraining the entire system is costly to maintain after the initial development. Recent work has proposed to augment the dialogue system with internet-search ability to generate more factual and updated responses in open-domain dialogues~\cite{komeili2021internet}. Future efforts in this direction will benefit social influence dialogue systems as well.

\section{Conclusions}
We introduced the category of social influence dialogue systems that aim to influence their partners through dialogue. We presented a survey of the recent prior work in this space, compiling datasets and methods across diverse application domains. We pointed out key limitations in existing methodologies and proposed promising directions for designing more sophisticated systems in the future. Our survey reveals that although substantial progress has been made, this is still an emerging research area. We hope our work inspires more dedicated interdisciplinary effort and discussion, which is necessary for making progress in this space.

\section{Broader Impact and Ethical Considerations}
Social influence is ubiquitous in everyday life. Research on how we use influence in all aspects of our lives spans a number of fields, including social psychology, communication, consumer behavior, behavioral change, and behavioral economics. This research has led to crucial findings about the strategies of social influence and how they impact our decision-making. Over the past few decades, research has accumulated and demonstrated the effectiveness of using various strategies across contexts and domains. Prominent examples include core principles of social influence by Cialdini from social psychology: reciprocity, commitment and consistency, social proof, liking and attractiveness, authority, and scarcity~\cite{cialdini2009influence}. Further, communication strategies used in persuasion and general social influence contexts include credibility appeals, two-sided argumentation, emotional tactics, and appeals to social norms, among others~\cite{cameron2009practitioner,o2015persuasion}.

First, the well-studied principles in social influence research can guide the development of effective dialogue systems with influence capabilities. In fact, many of the strategies found in the datasets developed for social influence tasks (Section \ref{sec:Social Influence Across Disciplines}) directly map to the principles laid out by Cialdini, for instance, credibility and emotional appeal in PersuasionForGood dataset~\cite{wang2019persuasion} and reciprocity observed in CaSiNo negotiation dataset~\cite{chawla2021casino}. Second, research in social influence dialogue systems provides novel datasets on human-human and human-machine communication, and therefore, holds a great potential to advance theories of human cognition and influence processes~\cite{gratch2015negotiation}. The datasets and subsequent analyses can further contribute new theoretical insights to social influence research.

Although dialogue systems have already been used in a number of applications involving chatbots and AI assistants, advancements in social influence dialogue systems can help to bridge the gap between our existing task definitions and a number of other real-world applications. For instance, realistic customer support interactions often involve active behaviors from both the support agent and the user where the agent uses social cues for improved customer satisfaction and retention, while the user attempts to address their queries. These settings naturally involve aspects of social influence, unlike traditional task-oriented definitions where the dialogue system plays a passive role to assist the human users. As discussed earlier, social influence dialogue systems can positively help to advance other areas as well. In therapy domain, these systems can assist in various psychological treatments such as by increasing the willingness to disclose~\cite{lucas2014s}. In pedagogy, they can help to make social skills training more accessible~\cite{johnson2019intelligent}.

While we think about these applications, it is crucial to also lay out proper ethical guidelines to avoid any misuse of these systems. Primary concerns are around the use of deception (e.g. in Diplomacy and other negotiation tasks), emotional appeals (e.g. in persuasion), and behavior change (e.g. in conversational recommendations).

To mitigate possible misuse scenarios or unintended harms, we now lay out a few ethical guidelines which also apply to dialogue research in general. First, rigorous attempts must be made to ensure that the data collection, design processes, and evaluations, strictly abide by the guidelines and regulations laid out by the relevant Institutional Review Board (IRB). Second, the research team needs to develop a thorough plan to monitor and understand the behaviors of the developed systems before deployment. This includes identifying the goals of the dialogue system, identifying potential toxic language use, and any discriminatory behaviors. Third, investment into improved data collection practices, along with explainable and controllable dialogue systems can help identify these issues early on and allow manipulation to avoid them. Fourth, we argue that transparency is the key. All stakeholders must be made aware of the goals and design objectives of the system, along with any known misbehaviors or potential risks. The users must also be informed of any data collected during the deployment phase. Lastly, we believe that continuous monitoring of dialogue systems is necessary to ensure that the system performs consistently and does not diverge to unexpected conditions that may incur offensive or discriminative actions. We hope that our work promotes a more systematic study of social influence dialogue systems, which in turn will help to tackle the ethical concerns in a more principled way.

\section{Limitations}
\label{sec:limitations}

\noindent\textbf{Literature Search}: We presented a survey of efforts in social influence dialogue systems. Although every attempt was made to provide the readers with a comprehensive overview of the research in this space, our work does not claim exhaustiveness in the covered literature and it is likely that we missed out on other relevant research in this space.

\noindent\textbf{Intention for influence}: The datasets and tasks covered in this literature review focus on scenarios where social influence is \textit{intentional by design}. However, social influence can also be \textit{unintentional}, that is, interactions between humans and machines can have unintended influence on the attitudes, behaviors, or feelings of the human user~\cite{gass2015social}. For instance, changes in topic preferences after interacting with a system on a variety of topics, or incorporating biases after interacting with a biased system. As we continue to make an unprecedented progress towards AI systems that interact with humans via natural means of communication, we must also take into account the unintended influence on the users of the underlying technology. We hope that our work motivates researchers to study these effects methodically in the future.

\section*{Acknowledgments}
We would like to thank our colleagues at the University of Southern California, Columbia University, and the University of California Davis, along with fellow researchers with whom we interacted at conferences, for all their comments and helpful discussions that have shaped this project. We also thank the anonymous reviewers for their valuable time and feedback. Our research was, in part, sponsored by the Army Research Office and was accomplished under Cooperative Agreement Number W911NF-20-2-0053. The views and conclusions contained in this document are those of the authors and should not be interpreted as representing the official policies, either expressed or implied, of the Army Research Office or the U.S. Government. The U.S. Government is authorized to reproduce and distribute reprints for Government purposes notwithstanding any copyright notation herein.



\bibliography{anthology,custom}
\bibliographystyle{acl_natbib}

\newpage

\pagebreak

\appendix

\section{Literature Compilation}
\label{sec:appendix-sources}
In this section, we provide details about how the literature was curated for our survey. We hope this helps the overall reproducibility and also guides similar studies in the future. We followed a simple two-stage process. First, we compiled the relevant datasets that capture various forms of social influence across diverse domains (presented in Section \ref{sec:Social Influence Across Disciplines}) and then, we compiled the techniques developed on these datasets (presented in Section \ref{sec:methods}).

\noindent\textbf{Step I - Datasets}: Our objective was to gather datasets that (by design) capture forms of social influence. We primarily focused on dialogue interactions but include the datasets based on transcripts from multimodal interactions as well. Given the large breadth of research in this space across a number of domains, our collection is not exhaustive but is rather restricted to the following sources. 

We surveyed the past $6$ years of *ACL conference proceedings. We then covered several online repositories of dialogue data to capture datasets published at other venues. This includes ParlAI\footnote{\url{https://github.com/facebookresearch/ParlAI}}, Huggingface\footnote{\url{https://huggingface.co/docs/datasets/index}}, NLP-Progress\footnote{\url{http://nlpprogress.com/english/dialogue.html}}, and Convokit\footnote{\url{https://convokit.cornell.edu/documentation/datasets.html}}. Further, we revisited several recent surveys in dialogue systems and Natural Language Generation (NLG) research~\cite{huang2020challenges,zhang2020recent,ni2021recent,duerr2021persuasive}. Datasets that were categorized as task-oriented or open-domain in these surveys but also contain some aspects of social influence have been included in our discussion. As discussed in Section \ref{sec:methods}, we also include the datasets that have not been directly used for designing dialogue systems but rather for various Natural Language Understanding (NLU) subtasks that can be crucial for the eventual development of dialogue systems in this space. Finally, we also reviewed the citation graphs of the collected papers from Google Scholar. Overall, we ended up with $22$ dataset papers, spanning $12$ publication venues, $4$ languages, and $7$ application domains.

\noindent\textbf{Step II - Methods}: Compiling the methodological progress was based on the models developed on the curated datasets. For this purpose, we simply reviewed the citations of all the dataset papers using Google Scholar.

\section{Datasets}
\label{sec:appendix-datasets}
A comprehensive list of the available datasets for investigating social influence in dialogues is provided in Table \ref{tab:corpora}. For each dataset, we mention the application domain, source, key statistics, as well as the available metadata and annotations apart from the conversation logs.

\begin{table*}[th!]
\centering
\scalebox{0.62}{
\begin{tabular}{lcP{3cm}p{4.5cm}p{6cm}}
\hline
\multicolumn{1}{c}{\textbf{Name (Citation)}} & \textbf{Domain} & \textbf{Source} & \multicolumn{1}{c}{\textbf{Key Statistics}} & \multicolumn{1}{c}{\textbf{Metadata \& Annotations}} \\ \hline
\makecell[l]{\textbf{STAC}\\\cite{asher2016discourse}} & Games & Crowdsource & \makecell[l]{Dialogues: 1081\\Turns/Dialogue: 8.5\\Tokens/Turn: 4.2} & \makecell[l]{Dialogue Acts;\\Discourse Structures}\\ \hline
\makecell[l]{\textbf{Diplomacy}\\\cite{peskov2020takes}} &  Games & Crowdsource & \makecell[l]{Games: 12\\Messages/Game: 1440.8\\Words/Message: 20.79$^+$} & \makecell[l]{Intended and perceived\\truthfulness; Participant demographics\\and self-assessment of lying abilities;\\Ground-truth betrayals}\\ \hline
\makecell[l]{\textbf{DinG}\\\cite{boritchev2021ding}} & Games & \makecell{University game\\ night logs} & \makecell[l]{Games: 10\\Turns/Game: 2357.5} & \makecell[l]{Annotated question-answer pairs}\\ \hline
\makecell[l]{\textbf{Tabletop}\\\cite{devault2015toward}} &  MIBT & \makecell{Face-to-face,\\Wizard-of-Oz} & \makecell[l]{Face-to-face Dialogues: 89\\Wizard-of-Oz Dialogues: 30} & \makecell[l]{Participant demographics; Subjective\\questionnaire-based assessment}\\ \hline
\makecell[l]{\textbf{DealOrNoDeal}\\\cite{lewis2017deal}} & MIBT & Crowdsource & \makecell[l]{Dialogues: 5808\\Turns/Dialogue: 6.6\\Tokens/Turn: 7.6} & \makecell{\textbf{---}} \\ \hline
\makecell[l]{\textbf{CaSiNo}\\\cite{chawla2021casino}} &  MIBT &Crowdsource & \makecell[l]{Dialogues: 1030\\Utterances/Dialogue: 11.6\\Tokens/Utterance: 22} & \makecell[l]{Participant demographics and\\personality traits; Outcome satisfaction;\\Partner perception;\\Strategy Annotations}\\ \hline
\makecell[l]{\textbf{JobInterview}\\\cite{yamaguchi2021dialogue}} &  MIBT & Crowdsource & \makecell[l]{Dialogues: 2639\\Turns/Dialogue: 12.7\\Words/Turn: 6.1} & Dialogue acts\\ \hline
\makecell[l]{\textbf{PersuasionforGood}\\\cite{wang2019persuasion}} & Social Good & Crowdsource & \makecell[l]{Dialogues: 1017\\Turns/Dialogue: 10.4\\Words/Turn: 19.4} & \makecell[l]{Participant sociodemographics,\\personality, and engagement in the\\conversation; Strategy annotations;\\Dialogue Acts}\\ \hline
\makecell[l]{\textbf{CraigslistBargain}\\\cite{he2018decoupling}} & E-commerce & Crowdsource & \makecell[l]{Dialogues: 6682\\Turns/Dialogue: 9.2\\Tokens/Turn: 15.5} & Dialogue Acts \\ \hline
\makecell[l]{\textbf{AntiScam}\\\cite{li2020end}} &  E-commerce & Crowdsource & \makecell[l]{Dialogues: 220\\Turns/Dialogue: 12.4\\Words/Turn: 11.1} & Dialogue Acts; Semantic Slots\\ \hline
\makecell[l]{\textbf{Motivational Interviewing}\\\cite{tanana2016comparison}} &  \makecell{Therapy \&\\Support} & \makecell{Psychotherapy\\session logs} & \makecell[l]{Sessions: 341\\Utterances/Session: 513.2\\Words/Utterance: 9.7} & Behavior annotations\\ \hline
\makecell[l]{\textbf{SMS Counseling}\\\cite{althoff2016large}} &  \makecell{Therapy \&\\Support} & SMS chat logs & \makecell[l]{Dialogues: 80,885\\Messages/Dialogue: 42.6$^*$\\Words/message: 19.2$^*$} & \makecell[l]{Post-conversation assessment for both\\the counselor (e.g. suicide risk, main\\issue etc.) and user (how they feel\\afterwards)}\\ \hline
\makecell[l]{\textbf{EmpatheticDialogues}\\\cite{rashkin2019towards}} & \makecell{Therapy \&\\Support} & Crowdsource & \makecell[l]{Dialogues: 24,850\\Utterances/Dialogue: 4.3\\Words/Utterance: 15.2} & \makecell{\textbf{---}}\\ \hline
\makecell[l]{\textbf{Hotline Counseling}\\\cite{demasi2019towards}} & \makecell{Therapy \&\\Support} & \makecell{Synthetic\\Transcripts} & \makecell[l]{Dialogues: 254\\Messages/Dialogue: 40-60} & \makecell[l]{Paraphrases by trained counselors}\\ \hline
\makecell[l]{\textbf{mPED}\\\cite{liang2021evaluation}} & \makecell{Therapy \&\\Support} & \makecell{Physical activity\\clinical trials} & \makecell[l]{Sessions: 107\\Turns/Session: 423.5\\Minutes/Session: 28.8} & \makecell[l]{Demographics; Physical activity related\\pre and post surveys; Strategy\\annotations}\\ \hline
\makecell[l]{\textbf{Congressional Debates}\\\cite{thomas2006get}} &  Argumentation & \makecell{U.S. Congressional\\ transcripts} & \makecell[l]{Debates: 53\\Speech segments/Debate: 72.8} & \makecell[l]{Ground-truth label with each speech\\segment for support/oppose of the\\proposed bill}\\ \hline
\makecell[l]{\textbf{Supreme Court}\\\cite{danescu2012echoes}} &  Argumentation & \makecell{Oyez.org\\transcripts} & \makecell[l]{Cases: 7700\\Utterances/Case: 220.8} & \makecell[l]{Case-related metadata such as key\\ dates, citation, parties involved, and\\voting results}\\ \hline
\makecell[l]{\textbf{DeCour}\\\cite{fornaciari2012decour}} & Argumentation & \makecell{Italian court\\hearings} & \makecell[l]{Hearings: 35\\Utterances/Hearing: 173.4\\Tokens/Utterance: 13.9$^\#$} & \makecell[l]{Metadata for testimonies like place,\\date, demographics; Hearing related\\details; Truthfulness annotations}\\ \hline
\makecell[l]{\textbf{ChangeMyView}\\\cite{tan2016winning}} & Argumentation & Reddit & \makecell[l]{Discussion Trees: 20,626\\Nodes/Tree: 61.1} & \makecell[l]{The original post with initial arguments\\and explicitly recognized successful\\persuasive arguments from the\\opposing side}\\ \hline
\makecell[l]{\textbf{DDO Debates}\\\cite{durmus2019corpus}} & Argumentation & debate.org logs& \makecell[l]{Debates: 78,376\\Messages/Debate: 7.7} & \makecell[l]{User demographics; Debate metadata \\like dates and category; Audience votes\\and comments}\\ \hline
\makecell[l]{\textbf{Court Debates}\\\cite{ji2020cross}} &  Argumentation & \makecell{China Court\\transcripts} & \makecell[l]{Dialogues: 260,190\\Utterances/Dialogue: 13.9} & \makecell{\textbf{---}}\\ \hline
\makecell[l]{\textbf{Target-Guided}\\\cite{tang2019target}} & Miscellaneous & Crowdsource & \makecell[l]{Dialogues: 9939\\Utterances/Dialogue: 11.4} & \makecell{\textbf{---}}\\ \hline
\end{tabular}}
\caption{\label{tab:corpora} Social Influence Dialogue Corpora, grouped by task domains and sorted by publishing year within a domain. All statistics of the form X/Y denote average numbers. MIBT: Multi-Issue Bargaining Task. $^*$Only computed for dialogues with additional survey responses, $^+$Only computed for the training data. $^\#$Only for Speaker utterances in front of the judge (doesn't include other members of the court). Note that not all datasets listed above have been directly used for designing end-to-end dialogue systems, but instead, these have enabled research into various sub-tasks and analyses that can eventually be useful for dialogue systems in this area. Please refer to Section \ref{sec:Social Influence Across Disciplines} in the main paper for a detailed discussion about these datasets and to Section \ref{sec:methods} for information about various methods developed using them.}
\end{table*}

\section{Five Stages for Designing Social Influence Dialogue Systems}
\label{sec:appendix-framework}
\begin{figure*}[th!]
\centering
 \includegraphics[width=\linewidth]{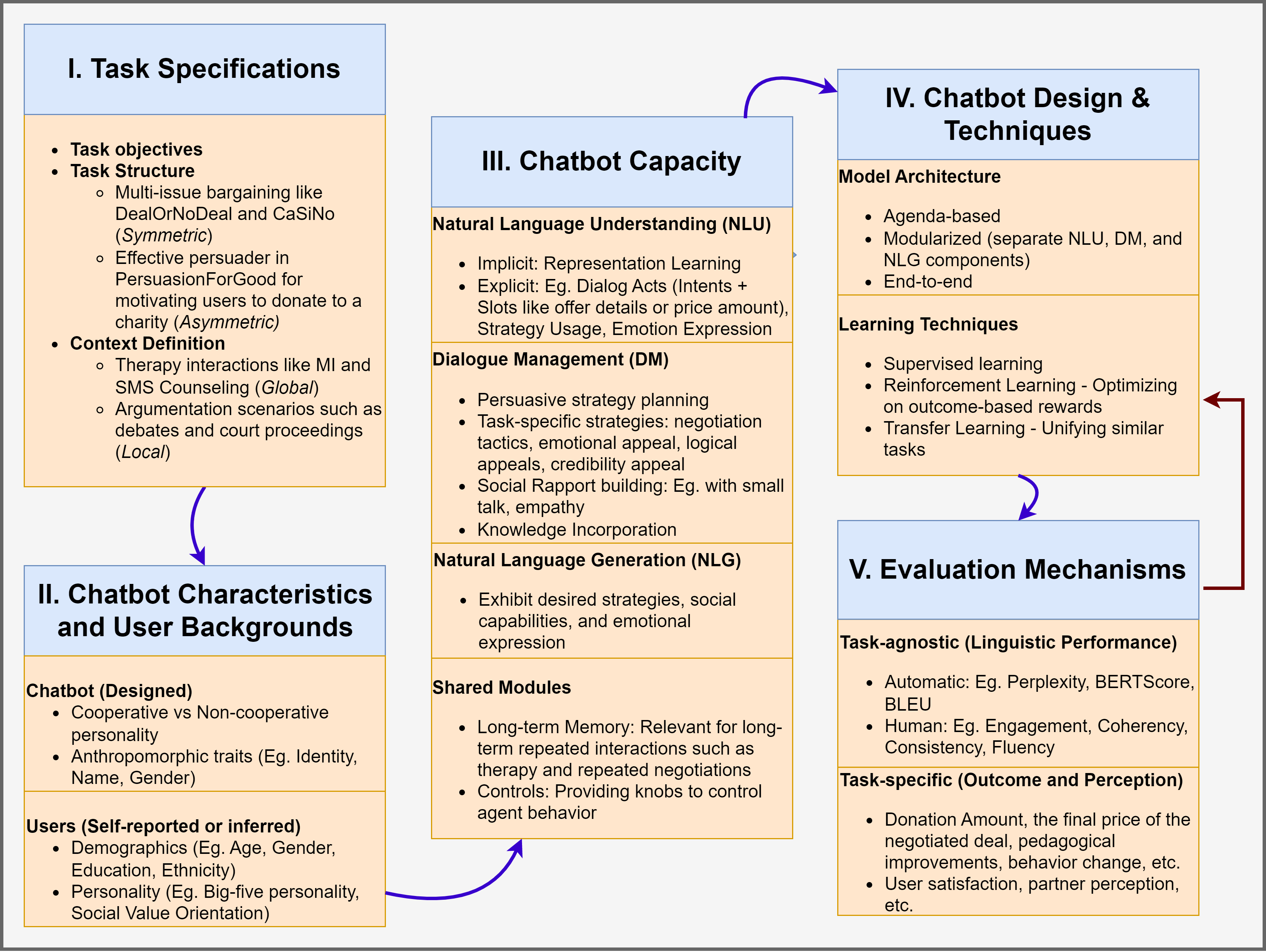}
\caption{A theoretical model for the development of dialogue systems for social influence tasks. Curved arrows represent forward relations and the straight arrow represents the feedback. \textbf{I. Task Specifications}: Key properties that define the task in consideration and are captured by the collected dataset, \textbf{II. Chatbot Characteristics and User Backgrounds}: Attributes for the agent design and target audience, \textbf{III. Chatbot Capacity}: The desirable capabilities of the system, \textbf{IV. Chatbot Design \& Techniques}: The modeling techniques to develop the dialogue system, and \textbf{V. Evaluation Mechanisms}: Metrics to evaluate system performance.}
\label{fig:chatbotdesign}
\end{figure*}

We develop a five-stage framework to summarize our recommendations for future work. These stages cover key decisions in the design of a dialogue system in this space, encouraging a holistic understanding of the system characteristics, target audience, underlying modeling techniques, and evaluation mechanisms. These steps are inspired by a behavior change model in healthcare research~\cite{zhang2020artificial}. We adapt this model to make it suitable for general social influence tasks in NLP. We present these steps in Figure \ref{fig:chatbotdesign}.

\end{document}